\documentclass[a4paper]{cas-dc}
\usepackage[square, comma, sort&compress, numbers]{natbib}

\newcommand{\M}{UTA-Sign}
\def\tsc#1{\csdef{#1}{\textsc{\lowercase{#1}}\xspace}}
\tsc{WGM}
\tsc{QE}

\begin{document}
\let\WriteBookmarks\relax
\def\floatpagepagefraction{1}
\def\textpagefraction{.001}


\shorttitle{\M: Unsupervised Thermal Video Augmentation via Event-Assisted Traffic Signage Sketching}   
\shortauthors{Yuqi Han et al}  

\title [mode = title]{\M: Unsupervised Thermal Video Augmentation via Event-Assisted Traffic Signage Sketching} 



%

\author[1]{Yuqi Han}[orcid=0000-0002-8097-1397]


\ead{yqhanSCST@dlut.edu.cn}



\affiliation[1]{organization={School of Computer Science and Technology, Dalian University of Technology},
            city={Dalian},
            postcode={116024}, 
            state={Liaoning},
            country={China}}

\affiliation[2]{organization={Deparment of Automation, Tsinghua University},
            city={Beijing},
            postcode={100084}, 
            country={China}}

\affiliation[3]{organization={School of Electrical and Electronic Engineering, Nanyang Technological University},
            postcode={639798}, 
            country={Singapore}}

\author[1]{Songqian Zhang}
\ead{2015@mail.dlut.edu.cn}

\author[1]{Weijian Su}

\ead{suwj@mail.dlut.edu.cn}

\author[3]{Ke Li}

\ead{LIKE0018@e.ntu.edu.sg}

\author[1]{Jiayu Yang}

\ead{jiayu_yang@mail.dlut.edu.cn}

\author[2]{Jinli Suo}

\cormark[1]

\ead{jlsuo@tsinghua.edu.cn}

\author[1]{Qiang Zhang}

\cormark[1]

\ead{zhangq@dlut.edu.cn}

\begin{abstract}
The thermal camera excels at perceiving outdoor environments under low-light conditions, making it ideal for applications such as nighttime autonomous driving and unmanned navigation. However, thermal cameras encounter challenges when capturing signage from objects made of similar materials, which can pose safety risks for accurately understanding semantics in autonomous driving systems.  
In contrast, the neuromorphic vision camera, also known as an event camera, detects changes in light intensity asynchronously and has proven effective in high-speed, low-light traffic environments. Recognizing the complementary characteristics of these two modalities, this paper proposes
\M, an \underline{u}nsupervised \underline{t}hermal-event video \underline{a}ugmentation for traffic \underline{sign}age in low-illumination environments, targeting elements such as license plates and roadblock indicators. 
To address the signage blind spots of thermal imaging and the non-uniform sampling of event cameras, we developed a dual-boosting mechanism that fuses thermal frames and event signals for consistent signage representation over time. The proposed method utilizes thermal frames to provide accurate motion cues as temporal references for aligning the uneven event signals. At the same time, event signals contribute subtle signage content to the raw thermal frames, enhancing the overall understanding of the environment.
The proposed method is validated on datasets collected from real-world scenarios, demonstrating superior quality in traffic signage sketching and improved detection accuracy at the perceptual level.
\end{abstract}


\begin{highlights}
\item  An unsupervised thermal-event augmentation method for traffic signage perception is proposed.

\item The proposed UTA-Sign leverages dual-modal complementarity for low-light signage sketching.

\item UTA-Sign delivers high-fidelity imagery and accurate detection under real-world traffic data.

\item The model extends perception ability and supports all-day driving with thermal cameras.


\end{highlights}

\begin{keywords}
 \sep thermal video sketching \sep neuromorphic-assisted augmentation \sep spatial-temporal correlation\sep unsupervised learning\sep low-illumination perception
\end{keywords}

\maketitle

\section{Introduction}\label{Int}
A fundamental goal of autonomous driving is to maintain safety under all lighting conditions. To meet this challenge, the perception module combines sensors and navigation systems to gather essential data about the traffic environment. This enables a comprehensive understanding of road conditions, which is crucial for informed decision-making. To achieve high-quality perception, it is essential not only to capture the environmental geometry but also to accurately interpret the details of the signage, such as road signs, obstacle indicators, and license plates. By grasping these key elements, vehicles can develop planning strategies designed specifically for the current traffic situation, resulting in better control and improved safety.

\begin{figure*}[t]
	\centering
    \vspace{2mm}
	\includegraphics[width=\linewidth]{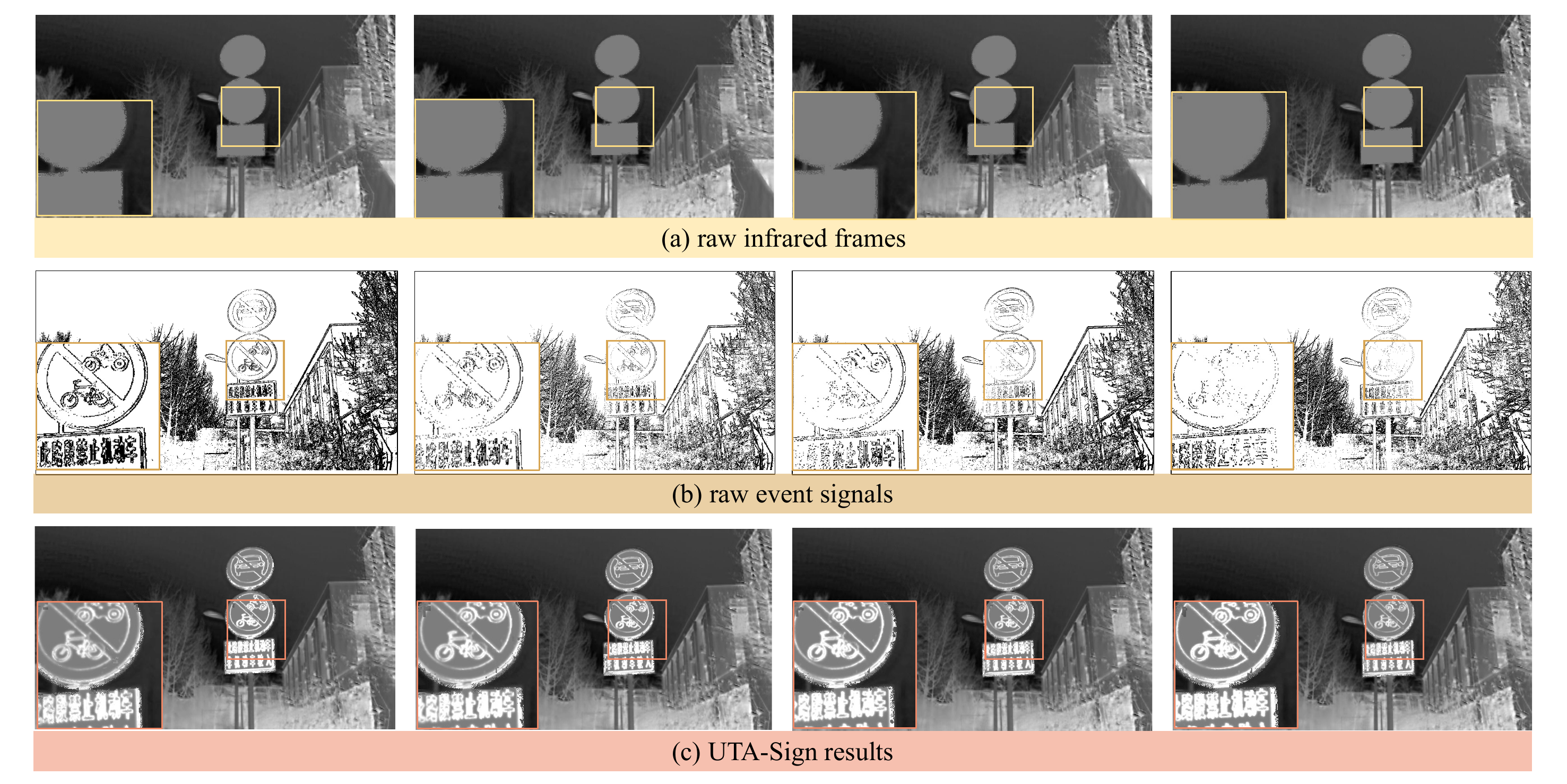}
	\caption{{Demonstration of frames of a typical driving scene captured by a thermal sensor (a) and event sensor (b), as well as the results reconstructed by our proposed approach (c).}}
    \vspace{2mm}
	\label{fig_tease}
\end{figure*}

The thermal camera works by detecting the radiation emitted from objects, enabling it to function effectively in total darkness without any need for additional light. Using temperature differences, these cameras clearly outline shapes and remain unaffected by surrounding light, such as ambient illumination or vehicle headlights, which ensures consistent visibility of thermal features. However, while thermal imaging excels in low-light settings and delineates edges, it falls short in capturing finer semantic details--for instance, traffic signs, as shown in Fig.~\ref{fig_tease}(a)—-because of the similarities in material and temperature. This shortcoming can impede a thorough understanding of the surroundings and pose risks to driving safety.

The fusion between thermal and optical sensors \cite{ma2019infrared,DU2025102859} has been widely investigated for complex perception environments and tasks. By integrating information from multiple sources, effective fusion solutions improve both accuracy and completeness.
Among optical sensors, neuromorphic cameras (events) stand out because they detect changes in light intensity rather than absolute values \cite{GHOSH2025102891,bardow2016simultaneous,scheerlinck2018continuous,munda2018real}, making them highly sensitive to illumination variations (Fig.~\ref{fig_tease}(b)). 
This system excels in delivering clear imagery even in dim lighting while effectively reducing motion blur, providing as a reliable method for enhancing thermal videos, bringing to light details about traffic signage that thermal sensors typically miss.

In this paper, we introduce \M, an \underline{u}nsupervised \underline{t}hermal-event \underline{a}ugmentation framework for \underline{sign}age sketching, recovering missing information in thermal frames, as shown in Fig.~\ref{fig_tease}. The proposed~\M~extends the feasibility of thermal sensors in low-light perception, further supporting applications such as autonomous driving.
Specifically, the \M~employs a dual-modal boosting fusion strategy, in which event signals and thermal frames collaboratively guide the fusion process. The fusion model leverages event signals to enrich the signage information in thermal videos while preserving their original content. Importantly, this approach operates in an unsupervised manner, without the need for costly labeled data, facilitating scalability and real-world deployment.

The framework integrates thermal and event signals in a frame-by-frame manner to create signage sketches. The Signage Information Sketching (SIS) module consists of modality-specific encoders to extract features, which are then fused across multiple scales before being decoded. To improve temporal consistency in thermal videos, the sketches are refined using a Temporal Consistency Correction (TCC) module, which aligns frames through deformable convolutions and captures temporal correlations using shifted window self-attention. The TCC module operates in a recurrent manner, with each output relying not only on the current thermal and event inputs but also on the preceding sketch frames. The model progressively refines the outputs, preserving essential features and ensuring temporally coherent and continuous results.

We independently create pseudo ground truths for the SIS and TCC module training. For the SIS module, we implement a masking mechanism to extract signage information from event signals. Specifically, the raw signals from both modalities act as prompts to segment signage information from the event signals, which is then mapped onto the thermal images. 
For training the TCC module, we use thermal signals to estimate the motion of the event camera. The aligned event signals, synchronized within a unified time frame, serve as the pseudo ground truth. This approach eliminates the need for feature matching across the temporal uneven event camera, a process that is particularly unreliable in a standalone operation. As a result, this constructed pseudo ground truth enhances the completeness of the semantics and improves the consistency of semantic sketching over time.

To validate the effectiveness of the proposed thermal sketching method, we conducted a series of comprehensive experiments. These experiments evaluated both qualitative visual quality and quantitative performance, and also assessed the capabilities of detecting traffic signs in low-illumination driving environments. Specifically, we compared the accuracy of object detection across three approaches: fusion of Thermal and RGB data, fusion of event camera and RGB data, and direct enhancement of RGB frames in low-light conditions. The experimental results demonstrate that the proposed thermal sketching method is more effective for all-day perception in autonomous driving.

\vspace{3mm}
The main contributions of this paper are listed as follows:
\begin{itemize}
\setlength{\itemsep}{0pt}
\setlength{\parskip}{0pt}
\setlength{\parsep}{0pt}
\item  This paper introduces a novel model that combines thermal and event cameras for the first time to reconstruct high-quality images of traffic signage in thermal videos, specifically for low-light autonomous driving. 

\item We employ an unsupervised training approach that leverages the strengths of thermal and event cameras for mutual enhancement: the former provides precise motion estimation for spatial-temporal compensation of events, while the latter contributes missing fine texture details in thermal videos.
\item We evaluated the proposed \M~comprehensively in real traffic scenarios, in terms of both imaging quality and downstream traffic-sign detection, and it demonstrates the optimal results under low-light driving conditions across various metrics.
\end{itemize}

\section{Related Work}\label{Rel}
Here we review the studies of low-light perception, including RGB–infrared (covering the thermal infrared spectrum) image fusion and event-based intensity image reconstruction, as well as object detection in low-light scenes.

\subsection{RGB-infrared Image Fusion}

RGB-infrared image fusion integrates complementary information from both modalities to enhance  visibility\cite{xu2020fusiondn,ma2022swinfusion,zhang2021sdnet,LI2024102147,Liang2022ECCV,wu2022difnet,xu2020u2fusion}. Current methods prioritize visual quality, lightweight architectures, and generalization. In terms of visual quality, FusionDN \cite{xu2020fusiondn} is proposed to balance quality and fidelity. The SwinFusion \cite{ma2022swinfusion} leverages cross-modal attention to enhance local complementarity and global consistency. SDNet \cite{zhang2021sdnet} integrates an adaptive decision block that modulates fusion based on local texture richness. CrossFuse \cite{LI2024102147} introduces a novel cross-attention mechanism to enhance complementary information and generate fused images through a two-stage training strategy. For lightweight optimization, a self-supervised decomposition method \cite{Liang2022ECCV} improves efficiency and robustness under limited data conditions. DIFNet \cite{wu2022difnet} proposes a parameterized pipeline that decouples extraction, fusion, and reconstruction to improve generalization. U2Fusion \cite{xu2020u2fusion} introduces adaptive information preservation for multiple fusion tasks.

Harsh illumination hinders semantic extraction from visible images, causing notable degradation. To address the issue, Tang et al. \cite{9834137} employs a dynamic Transformer to enhance salient structures in low-light scenes. Zhang et al. propose DAAF \cite{zhang2025daaf}, which decomposes images into frequency-domain components to separate illumination and reflection, thereby mitigating degradation. Leveraging large language models (LLMs), recent works incorporate semantic priors via text guidance to improve fusion quality \cite{yi2024text,zhang2025omnifuse}. Specifically, Text-IF \cite{yi2024text} utilizes semantic prompts to restore details in low-light regions. OmniFuse\cite{zhang2025omnifuse} introduces the Language driven model to image fusion task for eliminating degradation to improve the robustness of fusion results.  

The aforementioned methods demonstrate strong performance in RGB–infrared fusion, effectively enhancing image clarity and edge definition. However, they still face challenges in generating robust and high-fidelity semantic representations under severe illumination conditions.

\subsection{Event-based Intensity-Image Reconstruction}
Event cameras respond only to brightness changes, offering high temporal resolution.  A variational energy minimization framework \cite{bardow2016simultaneous} is developed to jointly estimate optical flow and reconstruct intensity images. Other approaches \cite{scheerlinck2018continuous,munda2018real} focus on continuous-time intensity estimation and real-time image reconstruction. Neural networks \cite{soltangholi2023intensity,lei2024many,geng2024event} are employed for event-assisted image reconstruction. 
Specifically, the research \cite{soltangholi2023intensity} enhances image quality given the sparse captured events.  On the basis, research \cite{lei2024many} models the optimal event number to reconstruct the high-quality intensity image. Research \cite{geng2024event} develops a multi-task collaborative framework, guiding the fusion results approximates the requirement of multi-tasks.
SPETNet \cite{zong2023single} is proposed to improve the speed and quality of intensity image reconstruction. Specifically, SPETNet exploits the temporal dimension by formulating the input as a tensor rather than a grid, enabling faster reconstruction.

Extending to color-scale reconstruction, the RGB-event fusion system is proposed to integrate event data with color cues, significantly expanding the application scope of event cameras. Diffusion models (DDPM) \cite{liang2023event} are applied to enhance image reconstruction quality. The edge information from the event signal is integrated into the DDPM model for sharp edge generation of the intensity image. Another method focuses on improving visual effects in reconstructed images \cite{li2024image}. Additionally, Guo et al. \cite{guo2023event} introduce the event camera super-resolution (EFSR-Net) network and design a coupled response block to improve the resolution and restore detailed textures. 

Although event cameras face challenges like excessive noise, they remain a key method for capturing fast texture information.  The event-based fusion solution could enhance perception ability and address downstream tasks.

\subsection{Object detection in Low-light Scenes}
Detection in low-light environments requires enhancing the contrast of key features in images and optimizing algorithms through techniques such as color space decomposition and spectral analysis to improve accuracy. 
The retinex theory, which models the relationship between object brightness, surface reflection, and ambient light, has inspired several enhancement-based approaches, including RetinexNet \cite{mahmood2024enhanced} and Gaussian-Retinex~\cite{9750714}.

Based on the illumination-reflection model, Yao et al.~\cite{yao2023traffic} enhance image illumination by converting RGB images to HSV and applying guided filtering for better object detection. Wu et al.~\cite{wu2023shadow} improve feature extraction, fusion, and classification to enhance detection under low light conditions. Additionally, Ren et al.~\cite{ren2023lightweight} combine coordinate attention and MBconv to increase both efficiency and accuracy in low-light detection.

Other low-light detection algorithms utilize the YOLO framework, as seen in works by Hui et al.~\cite{hui2024wsa}, Peng et al.~\cite{peng2024novel}, and Vinoth et al. \cite{vinoth2024lightweight}. Hui et al. \cite{hui2024wsa} effectively learn and perform object detection even with limited or incomplete labeled data by using a pre-enhanced image. Peng et al. \cite{peng2024novel} introduce NLE-YOLO, which reduces noise and effectively fuses features from various scales, thus improving detection accuracy for objects of different sizes in low-light conditions. Vinoth et al. \cite{vinoth2024lightweight} estimate the depth curve for each pixel, enhancing both detection speed and performance in dark scenarios. They also apply TensorRT optimization to reduce computational complexity and memory usage, enabling real-time detection in practical applications.

Currently, the available low-light detection methods are primarily designed for macroscopic objects, limiting their ability to capture fine-grained textures and semantically rich content. Accurately detecting such intricate details remains a challenging and unresolved issue in the field.

\section{Method}\label{Rel}

In this section, we first provide an overview of the proposed~\M~pipeline. We then detail the signage information sketching (SIS) and temporal consistency correction (TCC) modules, along with the construction of pseudo ground truth for each. The training loss is introduced subsequently. Finally, we briefly describe the preprocessing steps—including camera registration and semantic objective localization—as well as the training configuration.

\begin{figure*}[t]
	\centering
	\includegraphics[width=\linewidth]{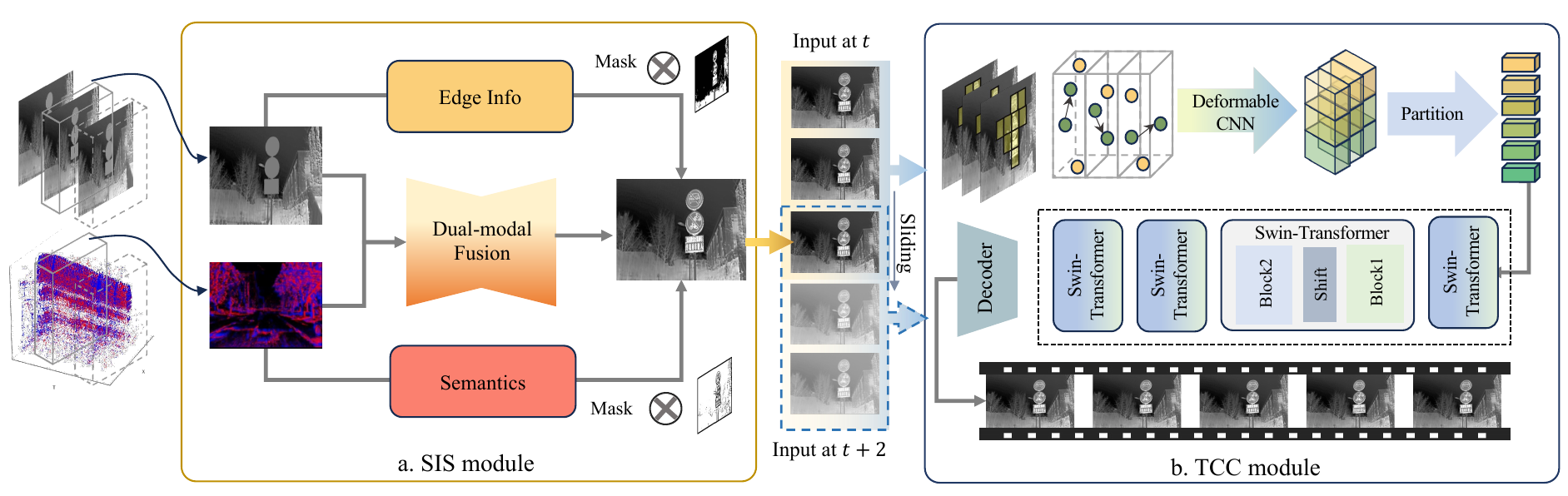}
	\caption{The pipeline of the proposed \M. We first send the registered infrared image and the event sequence to the SIS module for semantic sketching (a), and subsequently, adjacent frames enter the TCC module to exploit the temporal correlations and ultimately form a visually continuous video (b). }
	\label{fig_framework}
 \vspace{-2mm}
\end{figure*}

\subsection{Pipeline of \M~}

The pipeline of \M~is shown in Fig.~\ref{fig_framework}. The cameras are calibrated via the intrinsic and extrinsic parameters of two sensors, thereby enabling precise spatial alignment between the two modalities. Once the cameras are calibrated, the hard-triggered synchronization is employed to capture data simultaneously. 

The SIS is designed to map the signage information distinguished from event signals onto corresponding infrared frames, as shown in Fig.~\ref{fig_framework}(a). A hierarchical encoder–decoder architecture is employed to capture multiscale features and generate the signage sketches for infrared frames. The unsupervised learning strategy is adopted, wherein a mask guides supervision: regions within the mask reconstruct content from event signals, while regions outside the mask are supervised using the original infrared images.

The TCC module enhances spatiotemporal visual consistency of the sketches across multiple frames caused by the non-uniform density of event signals, as shown in Fig.~\ref{fig_framework}(b). Specifically, we introduce a deformable CNN to extract the temporally aligned feature across multiple frames, effectively modeling motion in the time domain. Later, the TCC extracts 3D features via shift-window spatiotemporal attention and infers the refined signage content of the target frame with a 2D vision transformer. The TCC module incrementally integrates information from preceding frames into the objective frame, enriching the semantic information while preserving visual consistency. Additionally, gradient loss and perceptual loss are incorporated to enhance the visual fidelity of the reconstructed sketches.

\begin{figure*}[t]
	\centering
	\includegraphics[width=\linewidth]{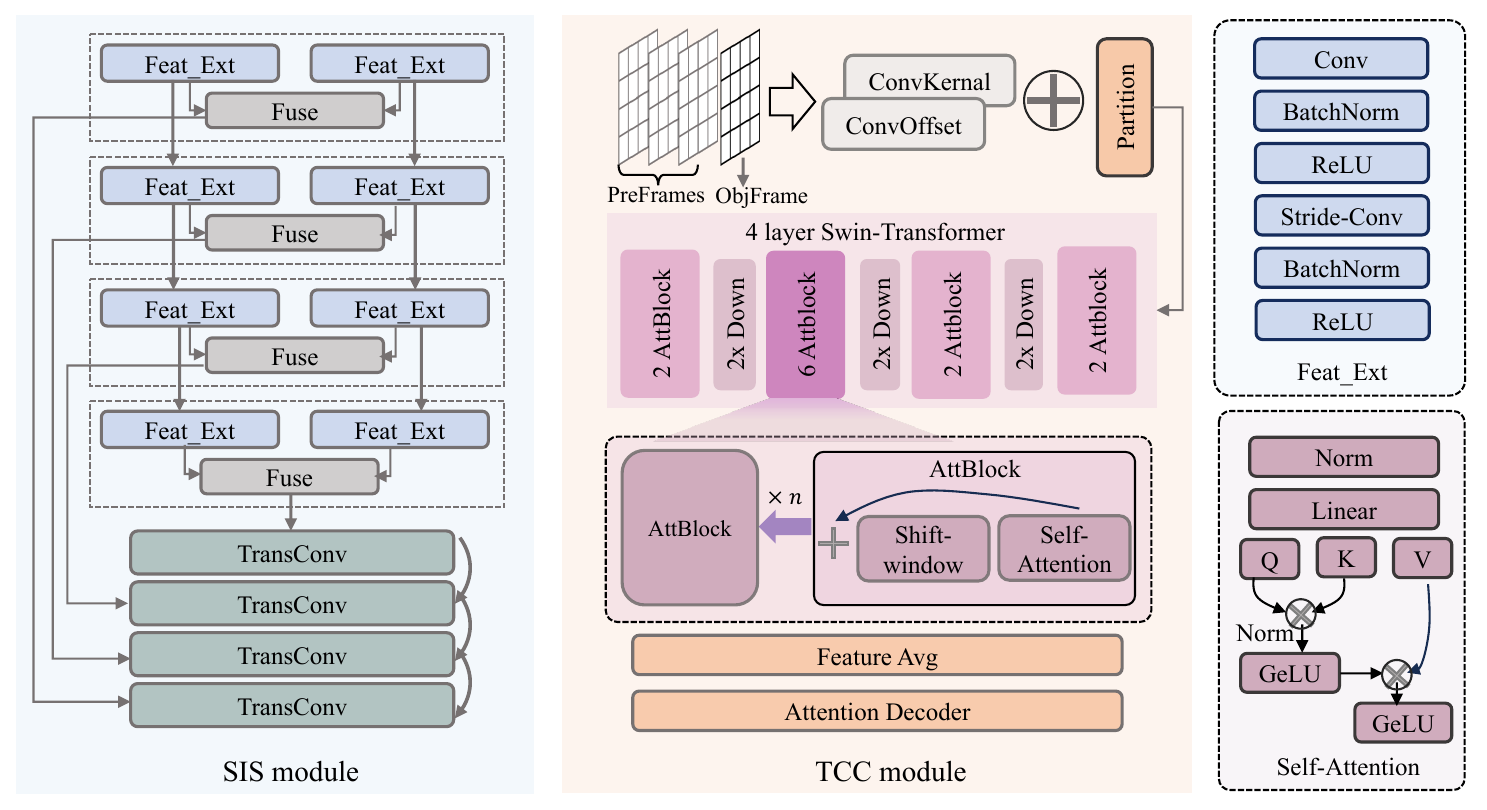}
	\caption{The network details of the key modules of the proposed \M.}
	\label{fig_frm}
 \vspace{-5mm}
\end{figure*}

\subsection{Signage Information Sketching}
The SIS module is designed to generate a coarse sketch of signage content by mapping event signals onto the thermal frames. As illustrated in the left part of Fig.~\ref{fig_frm}, the SIS adopts a dual-modal fusion framework based on a typical encoder–decoder architecture.

To simplify the feature extraction of continuous event signals, the event signals are partitioned into discrete event frames according to the frame rate of the thermal camera. Each event frame is independently paired with the corresponding thermal frame, thus the frame index $t$ is omitted. We denote the $I_{\text{EV}}$ and $I_{\text{EV}}$ as the thermal and event frames of each pair. The event encoder $E^*_{\text{EV}}$ and thermal encoder $E^*_{\text{IR}}$ individually construct the $K$ layers of feature extraction. The first layer of the feature is denoted as
\begin{equation}
F_{\text{EV}}^0 = E^*_{\text{EV}}(I_{\text{EV}}),\quad F_{\text{IR}}^0= E^*_{\text{IR}}(I_{\text{IR}}),
\end{equation}
and subsequent layers are progressively derived as
\begin{equation}
F_{\text{EV}}^k = E^*_{\text{EV}}(F_{\text{EV}}^{k-1}),\quad F_{\text{IR}}^k = E^*_{\text{IR}}(F_{\text{IR}}^{k-1}).
\end{equation}

The decoder constructs the coarse-to-fine feature concatenation. Given the event frame $I_{\text{EV}}$ and the thermal frame $I_{\text{IR}}$and each layer $k$ of the decoder is represented as
\begin{equation}
O_{k-1} = D_{\text{cov}}(UP(F_{\text{IR}}^k\odot F_{\text{EV}}^k)\odot O_{k}),
\end{equation}
where $D_{\text{cov}}(\cdot)$ indicates the transposed convolution, and $UP(\cdot)$ presents the up-sampling to the same size as $O_{k}$. The $\odot$ indicates concatenation between features. The final output $\hat I_{f} = O_{0}$ is the final fusion result of SIS.

\subsection{SIS Pseudo Ground Truth Construction}
The SIS module is designed to preserve both high-level signage details and low-level thermal cues, thereby enhancing the perceptual richness and expressiveness of the output sketches. To this end, a mask $M$ is constructed to distinguish semantic regions within the scene and guide the fusion of event signals and thermal images. Specifically, we assume that $S$ signage region with semantics are present. Then we have the union of the regions, denoted as $EV = EV_{0}\cup EV_{1}\cup ...\cup EV_{S}$. The mask $M$ is defined as 
\begin{equation}
 M(i,j)=\left\{
\begin{split}
&1 \quad \left\{i,j\right\} \in EV,\\
&0 \quad \text{Otherwise.}
\end{split}
\right.
\end{equation}
where $i,j$ indicate the indices of pixels. 
According to the definition of the mask, we construct a pseudo ground truth, denoted as 
\begin{equation}
{G}_{\text{SIS}}= M\times  I_{\text{EV}} + (1-M)\times I_{\text{IR}}.
\end{equation}

For each event–thermal frame pair, the mask and pseudo ground truths are constructed independently, thus we omit the temporal index.
\subsection{Temporal Consistency Correction}

Since the SIS module primarily focuses on short-term information, it overlooks the temporal consistency and smoothness of the sketches across consecutive frames.  
The asynchronous nature of event cameras may lead to signal missing and unpredictable noise. As a result, semantic sketches derived from event and thermal image fusion often exhibit inconsistent visual quality, thereby hindering high-fidelity real-world traffic perception.
To address this issue, we introduce the TCC module to enhance temporal continuity.

As illustrated in Fig.~\ref{fig_frm}, the input to TCC comprises $N$ frames: the SIS output of the target frame $\hat I_{f}(t)$ and its $N-1$ preceding frames. These frames are used to generate the final result $I^*_{f}(t)$ at time $t$. For clarity, and without loss of generality, we define  $V = \left\{\hat I_{f}(t-N+1), \ldots, \hat I_{f}(t-1),\hat I_{f}(t)\right\}$, and omit the explicit time index $t$ in the following discussion.

The deformable CNN is performed to accommodate complicated object motion, which aligns the similar features described in different frames, effectively enhancing temporal coherence and structural consistency. Unlike standard convolutions with fixed sampling grids, the deformable CNN adaptively samples from irregular positions, enabling more effective capture of structural edges and fine details within the volume. We define the Deformable CNN $D$ associate with an offset estimation convolution network $D_o$, the output feature of the deformable CNN is
\begin{equation}
 \mathbf{F_D} = D(V,D_o(V)),
\end{equation}
where $D_o(V)$ denotes learned offsets of sampling positions.

Later, a 4-layer swin-transformer is employed to learn deformation features across multiple feature scales. 
The input of each layer $l$ is denoted as $\mathbf{F}^{swin}_{l}$, where $\mathbf{F}^{swin}_{1} =  \mathbf{F_D}$ for the first layer. We denote the swin-transformer module of layer $l$ as $S_l$, and the feature downsampling as $d_l$.  The formulation is presented as
\begin{equation}
\mathbf{F}^{swin}_{l+1} = D_{l}(S_{l}(\mathbf{F}^{swin}_{l}),
\end{equation}
where $\mathbf{F}^{swin}_{out}(t) = \mathbf{F}^{swin}_{4}$ for the final layer.

The swin-transformer layer $S_l$ consists of multiple attention blocks. In layer $l$, we define $B$ attention blocks and the $b$-th is represented as $\mathcal{B}_{l,b}$. The $S_l$ is defined as
\begin{equation}
    S_{l}(\mathbf{F}^{swin}_{l}) \equiv \odot_{b=1}^{B} \mathcal{B}_{l,b}(\mathbf{F}^{swin}_{l}),
\end{equation}
where $\odot$ indicates the cascade calculation. In impletation, layers 1, 2, and 4 each contain 2 blocks, whereas layer 3 consists of 6 attention blocks for fully exploring the features.

Each attention block generally consists of a self-attention mechanism and a shift-window strategy, except for the final block, which does not employ window shifting. 
The self-attention mechanism $\text{Att}_{3d}$ calculates a weighted sum of all elements, where the weights are based on the similarity between the query and the keys. The shifted-window strategy $\text{Roll}(F,\textbf{s})$ performs the long-range feature interaction, where $\textbf{s}$ indicates the range size. We define the input and output features are $F^b_{in}$ and $F^b_{out}$, respectively. Given the window size $ M_D \times M_H \times M_W$, the process of attention block $\mathcal{B}_{l,b}$ includes
\begin{equation}
\begin{split}
F^b_{out}&= \text{Att}_{3d}(\text{LayerNorm}(F^b_{in})),\\
F^b_{out} &=\text{Roll}(F^b_{out}, \mathbf{s}) +  F^b_{out}, \\
\text{where}&
\begin{cases}
\mathbf{s} = (0,0,0),& b\text{ is final block}\\
\mathbf{s} = \left(\frac{M_D}{2}, \frac{M_H}{2}, \frac{M_W}{2}\right),& \text{otherwise.}
\end{cases}
\end{split}
\label{block}
\end{equation}

After the four layers, the feature is decoded by a 2D transformer $\text{Att}_{2d}$ for the objective frame, which is presented as
\begin{equation}
    I^*_{f} = \text{Att}_{2d}(\text{Avg}(\mathbf{F}^{swin}_{out}))
\end{equation}
where $\text{Avg}()$ averages of features to the initial input size.

The pipelined strategy is employed where each frame assimilates complementary cues from preceding neighbors. The preceding frames guide the sketching of the current frame, enabling future online inference for real-world implementation.

\begin{figure}[t]
	\centering
	\includegraphics[width=\linewidth]{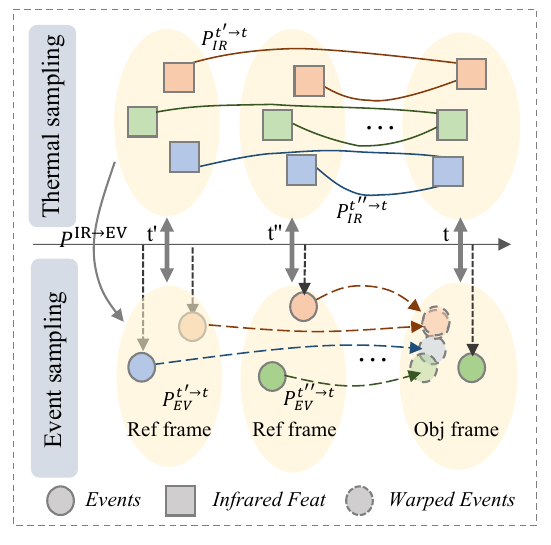}
	\caption{The construction principle of the TCC pseudo ground truth that leverages motion estimation from thermal signals to achieve temporal alignments with the events. }
	\label{fig_tcc}
\end{figure}

\subsection{TCC Pseudo Ground Truth Construction}

In this subsection, we construct pseudo ground truth with consistent sketches to guide the training of the TCC module. We estimate the camera motion and warp the temporally asynchronous event signals to a determined time instant, thereby completing the semantics. Moreover, during a given time interval, by warping the asynchronously captured event signals to respective instants, yielding a continuous and view-consistent representation.

The asynchronous captured event signals hinder relative pose estimation via feature registration of event frames. Given that the relative pose between the thermal camera and the event camera remains constant,  the event camera’s pose change is derived from the thermal camera’s pose change, as shown in Fig.~\ref{fig_tcc}. The pseudo ground truth is constructed over $T$ frames based on the sampling rate of the thermal camera. We denote the $T$ thermal frames as $I_{\text{IR}}^1, I_{\text{IR}}^2, \dots, I_{\text{IR}}^{T}$. The corresponding partitioned event frames are denoted as $I_{\text{EV}}^1, I_{\text{EV}}^2, \dots, I_{\text{EV}}^{T}$. The relative pose of the thermal camera from $t'$ to $t$ is denoted as
 $P_{\text{IR}}^{t'\rightarrow{} t}$. Meanwhile, the relative pose from the thermal camera to event camera is denoted as $P^{\text{IR}\rightarrow{}\text{EV}}$.
We denote the pose of the event camera from $t'$ to $t$  as $P_{\text{EV}}^{t'\rightarrow{}t}$, which is derived as
\begin{equation}
P_{\text{EV}}^{t'\rightarrow{}t} = P^{\text{IR}\rightarrow{}\text{EV}}P_{\text{IR}}^{t'\rightarrow{} t} (P^{\text{IR}\rightarrow{}\text{EV}})^{-1}.
\end{equation}

Given the pose $P_{\text{EV}}^{t'\rightarrow{}t} $ and the image $I_{\text{EV}}^t$, the warped image is denoted as $W(P_{\text{EV}}^{t'\rightarrow{}t},I_{\text{EV}}^t)$. We consider $t$ as the objective frame and others are reference frame, the pseudo ground truth of frame $t$ is derived as
\begin{equation}
    G_{\text{TCC}}(t) = \mathbf{1}(\sum_{t'=1}^{T}W(P_{\text{EV}}^{t'\rightarrow{}t},I_{\text{EV}}^{t'})+ I_{\text{EV}}^{t})
\end{equation}
where  $\mathbf{1}$ is an indicator function. Pixels are binarized by setting them to 1 if they exceed a predefined threshold, and to 0 otherwise. The threshold is set to $\frac{T}{2}$ in the experiment.

To mitigate noise inherent in event data, we perform denoising post-warping by assuming that noise is independently distributed in the spatiotemporal domain. Isolated event points lacking local spatiotemporal support are identified as noise and removed.

\subsection{Loss Function}

The SIS module and TCC module are trained jointly while being constrained to independent loss functions.
We construct an $\mathcal{L}_1$ loss to enable the output of SIS to approximate the spatial pseudo ground truth as closely as possible. Given the SIS output $\hat I_{f}(t)$ at time $t$, the SIS loss is
\begin{equation}
    L_{\text{SIS}}(t) = |\hat I_{f}(t),{G}_{\text{SIS}}(t)|_1.
\end{equation}

TCC loss enforces temporal consistency by requiring the current frame’s output to resemble both the current input and the corresponding representations of preceding frames.
Based on the output $I^*(t)$ at $t$, the TCC loss function is defined as
\begin{equation}
L_{\text{TCC}}(t) = |I^*_f(t) - G_{\text{TCC}}(t)|_1.
\end{equation}

In addition, we introduce perceptual loss $L_{per}$ for the signage region to further constrain the fusion results. Comparing perceptual similarities in the high-level feature space makes the fused reconstructed semantics visually more natural.
Gradient loss $L_{G}$ is introduced to avoid the distortion of high-frequency information by averaging the details of the image. We introduce the Laplacian operator to extract the edges of the semantic part and enhance them at the semantic edges. In summary, the total loss is written as
\begin{equation}
    Loss = \sum_{t=1}^T\big(L_{\text{SIS}}(t) + L_{\text{TCC}}(t) + L_{per}(t) + L_{G}(t)\big).
\end{equation}


Based on the loss function, we construct the end-to-end training integrating the SIS loss and the TCC loss. The SIS loss leverages the network to generate the clear sketched semantics, and the TCC loss constrains the temporal consistency of visual perception.

\subsection{Preprocessing and Training Details}
The registration of dual-camera systems involves aligning the cameras in both time and space. To achieve time-domain alignment, we generate an external signal that continuously outputs waveforms at predefined intervals. For spatial pose estimation, we utilize an thermal-recognizable checkerboard calibration board for corner detection and camera pose estimation.

It is important to note that the two cameras are fixed side by side. Due to variations in their fields of view and resolutions, the captured images cannot be directly aligned at the pixel level. To resolve this issue, we perform camera registration by aligning the grayscale frames from the DAVIS 346 sensor with the corresponding thermal frames. This is achieved by extracting and matching distinctive features from both image modalities, which allows us to accurately register the camera positions from the two perspectives.

\begin{figure}[t]
	\centering	\includegraphics[width=\linewidth]{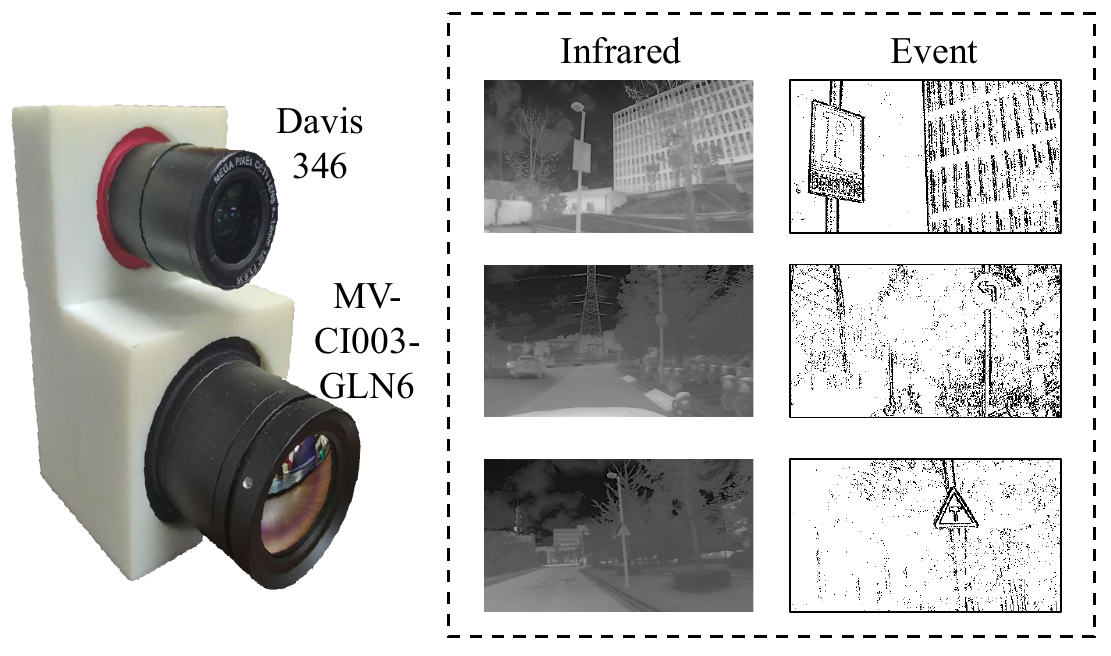}
	\caption{Real-world dual-modality acquisition system and representative outdoor captured data.}
	\label{datacapdevice}
 \vspace{-5mm}
\end{figure}

We implemented the network training and inference with PyTorch on NVIDIA RTX 4090 under Ubuntu 20.04. In the training stage, we adopted Adam optimizer with $\beta_1 = 0.9, \beta_2 = 0.9$, and $\beta_3 = 0.95$. The learning rate is initialized to $5e-4$ and gradually decayed to $1e-6$ using the linear reduction. We randomly cropped 1/4 of the original image and employed random flipping and rotation as data augmentation tricks to enhance the image representation. The batch size was set to 16. We conducted 300 epochs until convergence. The total parameter size of the proposed \M~is 38.27M and the FLOPs is 218.05G for the input size of (448, 448).

\begin{figure*}[t]
	\centering
	\includegraphics[width=1.0\linewidth]{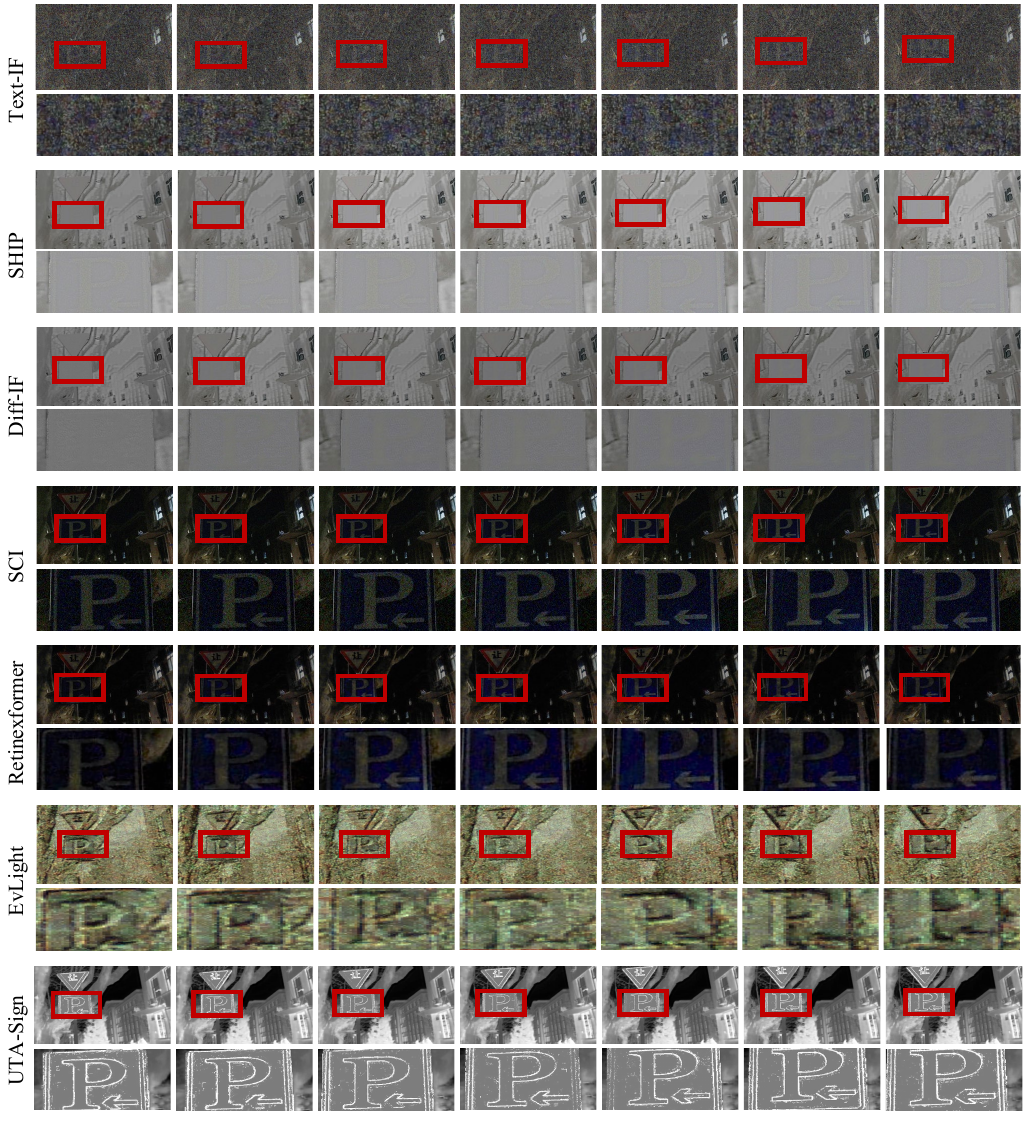}
	\caption{Visual comparison against existing methods on 7 consecutive frames from a simulated video.}
	\label{result1}
 \vspace{-3mm}
\end{figure*}

\section{Experiment Results}
In this section, we thoroughly evaluate the performance of our method, referred to as \M. We start by explaining the process of dataset construction. Next, we introduce the evaluation metrics and the baseline comparisons. We then present both qualitative and quantitative results obtained from real-world scenarios to demonstrate the effectiveness of \M. Additionally, we perform objective detection as a downstream task to showcase the system's capabilities in low-illumination conditions. Finally, we conduct ablation studies to analyze the various modules.

\subsection{Dataset Construction}
We curated a comprehensive dataset encompassing a wide range of low-light night-time traffic scenarios. The dataset comprises 34 distinct scenes and a total of 1,530 data groups, with each group consisting of 7 consecutive frames. It is structured into two components: real-world collected data and simulated data. 

\vspace{2mm}
\noindent{\textbf{Real-world Data.}\quad}
The real-world data is collected under nighttime conditions to showcase the raw data in low light. The acquisition system integrates a thermal camera and an event camera, as illustrated in Fig.~\ref{datacapdevice}. The thermal camera used is the HikVision industrial long-wave thermal model MV-CI003-GLN6, which operates within the 8–14 $\mu$m spectral range. It offers a resolution of 640 × 512 pixels and supports a frame rate of 50 frames per second. The event camera, DAVIS 346, can produce up to 12 million events in 1 $\mu$s, with a spatial resolution of 346 × 260 pixels.

\vspace{2mm}
\noindent{\textbf{Simulated Data.}\quad} 
The simulated data is derived from RGB images captured in real-world low-light scenes. We then apply the method proposed in the sRGB project \cite{lee-2023-edgemultiRGB2TIR} to convert these RGB images into approximate thermal representations.  Additionally, event data are synthesized following the methodology outlined in rpd\_vid2e \cite{gehrig2020video}. For each reference RGB frame, we aggregate events within a 30-ms temporal window to construct a single event frame, ensuring temporal alignment between each pair of captured data.

\begin{figure*}[t]
	\centering	
    \vspace{2mm}    \includegraphics[width=1.0\linewidth]{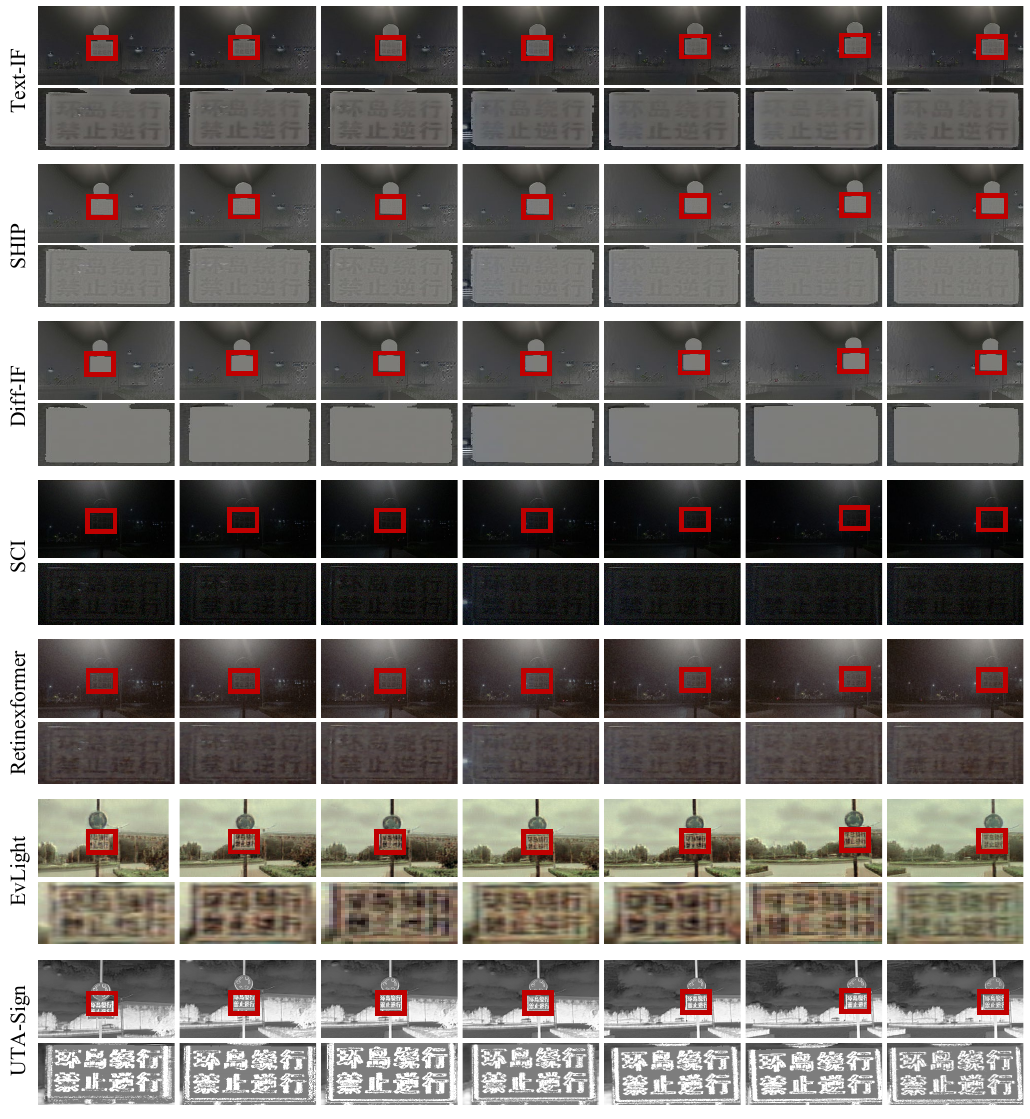}
	\caption{Visual comparison of our results against existing methods on 7 consecutive frames from a real dark video.}
	\label{result2}
 \vspace{-5mm}
\end{figure*}

\subsection{Metrics and Baselines}

\vspace{2mm}
\noindent {\textbf{Metrics.}\quad} We employ NIQE, PI, EN, SD, CNNIQA as the no-reference image quality metrics to evaluate the performance of fusion algorithms, all the indication of the metrics is thoroughly described in research \cite{ma2019infrared}. These metrics investigate whether the fused images satisfy human perceptual requirements and show clear edges, thereby validating which thermal-event fusion scheme satisfies human visual interpretation of the scene.

NIQE evaluates image quality by measuring deviations from natural scene statistics, where lower scores indicate higher perceptual quality.  PI is defined as a composite metric with lower values signifying better visual fidelity. EN quantifies the amount of information or detail in an image, with higher values indicating greater richness. SD measures the dispersion of pixel intensities and serves as an indicator of contrast. CNNIQA implicitly establishes the relationship between image content and perceived quality through a neural network, providing a robust and accurate assessment of feature's clarity.

\vspace{2mm}
\noindent {\textbf{Baselines.}\quad} We utilize six baseline algorithms to assess the semantic quality of images captured in low-light conditions. 
Given the absence of direct fusion for thermal-event signals, several RGB imaging–related methods are investigated to validate the~\M's effectiveness in low-light scenarios.
Among these algorithms, Text-IF \cite{yi2024text}, SHIP \cite{10655996}, and Diff-IF \cite{YI2024102450} are prominent methods for fusing visible and thermal images. Text-IF employs a text-guided image fusion technique that establishes an interactive framework to improve perceptual quality, especially in degraded scenarios. SHIP enhances the model's ability to collaborate across multiple modalities, integrating both global and fine-grained features across spatial and channel dimensions. Diff-IF introduces a difference mapping that effectively reduces training instability and prevents pattern collapse in the fusion results.

On the other hand, SCI \cite{ma2022toward} and Retinexformer \cite{cai2023retinexformer} focus on enhancing RGB images captured in low-light environments. SCI is an unsupervised method that employs a cascaded approach, wherein each stage independently calibrates and refines image quality. Retinexformer features an illumination-guided Transformer designed to estimate lighting conditions and facilitate interactions across different regions of an image to capture long-range correlations.

Finally, EvLight \cite{liang2024towards} integrates information from RGB and event-based cameras to enhance visual quality. Specifically, EvLight constructs a comprehensive dataset of paired RGB and event-based images, enabling the reconstruction of low-light visible images that closely approximate the original RGB images captured under natural lighting conditions.

\begin{table*}[t]
\centering
\caption{Quantitative comparisons of different methods.}
\begin{tabular}{cccccccc}
\Xhline{1pt}
~~~Metrics~~~ & ~~~Text-IF~~~ & ~~~SHIP~~~ & ~~~Diff-IF~~~ & ~~~~~SCI~~~~~ & Retinexformer & ~~~~EvLight~~~~ & ~~~\M~~~ \\
\Xhline{1pt}
NIQE$_\downarrow$ & 5.724 & 4.846 & 5.335 & 13.245 & \textcolor{blue}{\textbf{4.135}} & 7.915 & \textcolor{red}{\textbf{4.117}} \\
PI$_\downarrow$ & 4.654 & 4.243 & 4.538 & 8.288 & \textcolor{blue}{\textbf{3.887}} & 5.350 & \textcolor{red}{\textbf{3.287}} \\
EN$_\uparrow$ & 5.549 & 5.762 & 5.659 & 5.379 & 6.471 & \textcolor{blue}{\textbf{7.119}} & \textcolor{red}{\textbf{7.344}} \\
SD$_\uparrow$ & 15.547 & 15.517 & 15.612 & 15.647 & 25.375 & \textcolor{blue}{\textbf{52.998}} & \textcolor{red}{\textbf{66.080}} \\
CNNIQA$_\uparrow$ & 0.576 & 0.591 & \textcolor{blue}{\textbf{0.665}} & 0.419 & 0.380 & 0.417 & \textcolor{red}{\textbf{0.676}} \\
\Xhline{1pt}
\label{quan}
\end{tabular}
\end{table*}

\subsection{Qualitative Results}
We conducted qualitative experiments using two types of scenarios: one based on simulated data and the other on real-world collected data. The results for the simulated data are illustrated in Fig.~\ref{result1}, while the outcomes based on real-world collected data are presented in Fig.~\ref{result2}. In each scenario, we present 7 consecutive frames to evaluate temporal continuity and visual quality for each method.

Fig.~\ref{result1} displays the semantic reconstruction of a parking sign. Under extremely low light conditions, the fusion of RGB and thermal images (Text-IF, SHIP, Diff-IF) makes it challenging to identify the semantics clearly. The results of low-light enhancement methods (SCI, Retinexformer) still appear dark and require careful examination to discern any improvement. Although EvLight enhances brightness in ultra-low light conditions, its reconstructed results exhibit higher noise levels compared to our method, \M. This increased noise results from EvLight's failure to align events in the time domain; using raw events for scene reconstruction leads to inconsistent timing of signals.

Fig.~\ref{result2} showcases the semantic reconstruction of a Chinese traffic sign that reads ``Roundabout — No Wrong-Way Driving Allowed,'' based on real-world data. In this scenario, \M~displays the clearest semantic information. In low-light conditions, thermal images lack surface semantic details, and RGB images provide limited usefulness. As a result, the RGB-thermal fusion outcomes closely resemble the thermal imagery alone. Direct enhancement of RGB images results in only marginal improvements in clarity and demonstrates poor adaptability. While the RGB and event fusion approach enhances overall brightness, it ultimately falls short in fidelity and consistency.

In conclusion, Figs.~\ref{result1} and \ref{result2} illustrate that when RGB fails to capture semantic information, all RGB-related solutions also fail. In contrast, event cameras effectively capture lighting changes and compensate for the semantic limitations, making them well-suited for complementary fusion with thermal cameras in low-light scenarios.

\subsection{Quantitative Results}
This subsection presents quantitative results from perceptual metrics and the detection accuracy of traffic signs to verify the effectiveness of our approach.

\vspace{2mm}
\noindent \textbf{Perceptual Metrics.\quad}
The \M~represents the initial method aimed at reconstructing traffic signage by integrating thermal and event signals. In low-illumination scenarios, the lack of ground-truth references for texture renders conventional pixel-level evaluation impractical. Therefore, we utilize perceptual evaluation metrics to assess the visual quality of the reconstructed images.

Tab.~\ref{quan} presents the quantitative results. The method \M~outperforms other approaches across multiple evaluation metrics, indicating a significant advantage in perceptual quality. Retinexformer achieves the second-best performance in the NIQE and PI metrics, while Evlight ranks second in terms of EN and SD. Diff-IF shows relatively strong performance on the CNNIQA metric. Retinexformer primarily performs global brightness adjustments and denoising, which enhances the image's natural appearance. Although Evlight improves texture richness and contrast metrics, it suffers from excessive high-frequency noise, which negatively impacts its visual quality.

\begin{figure*}[t]
	\centering
	\includegraphics[width=1.0\linewidth]{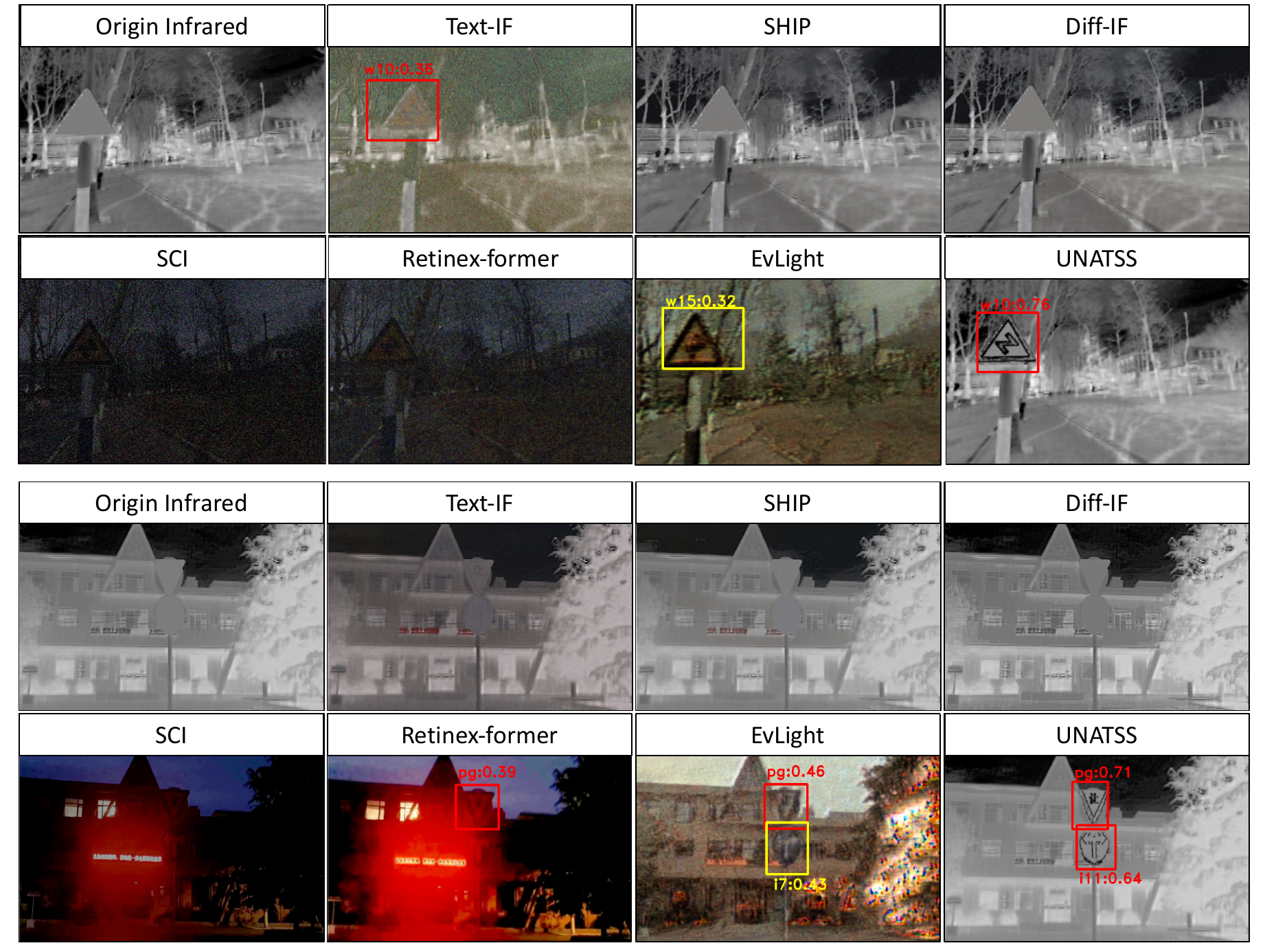}
	\caption{The comparison of the visual results produced by our approach and existing methods, in terms of the detection accuracy of the traffic signs.}
	\label{detectionresult}
 \vspace{-5mm}
\end{figure*}

\vspace{2mm}
\noindent \textbf{Detection Accuracy.\quad} 
Since object detection is a crucial task in autonomous driving, we evaluate \M~and the baseline models with respect to their accuracy in detecting traffic signs, as shown in Fig.~\ref{detectionresult}. The detection module is implemented by training YOLOv9 on the TT100K dataset. To illustrate detection performance, we present two representative scenes, in which red bounding boxes indicate correct detections and yellow bounding boxes indicate incorrect ones. In the first scene, Text-IF, EvLight, and the proposed \M~can successfully identify the traffic sign; however, EvLight produces an incorrect detection result. Both Text-IF and \M~effectively localize and accurately classify the semantic content, with \M~exhibiting superior accuracy.

The second scene in Fig.~\ref{detectionresult} contains two traffic signs. In this case, \M~successfully identifies both signs. EvLight localizes both signs but accurately detects only one of them, while Retinexformer correctly detects and classifies a single sign. Although the image enhancement and fusion methods effectively reconstruct structural elements such as buildings, they struggle with the fine-grained signage details. The proposed~\M~achieves reliable semantic sketching to support downstream tasks, enabling the implementation in practical scenes.

\subsection{Ablation Studies}
In this subsection, we evaluate the impact of the temporal consistency constraint module and the loss function on the \M~model separately.

\begin{figure*}[t]
	\centering
	\includegraphics[width=1.0\linewidth]{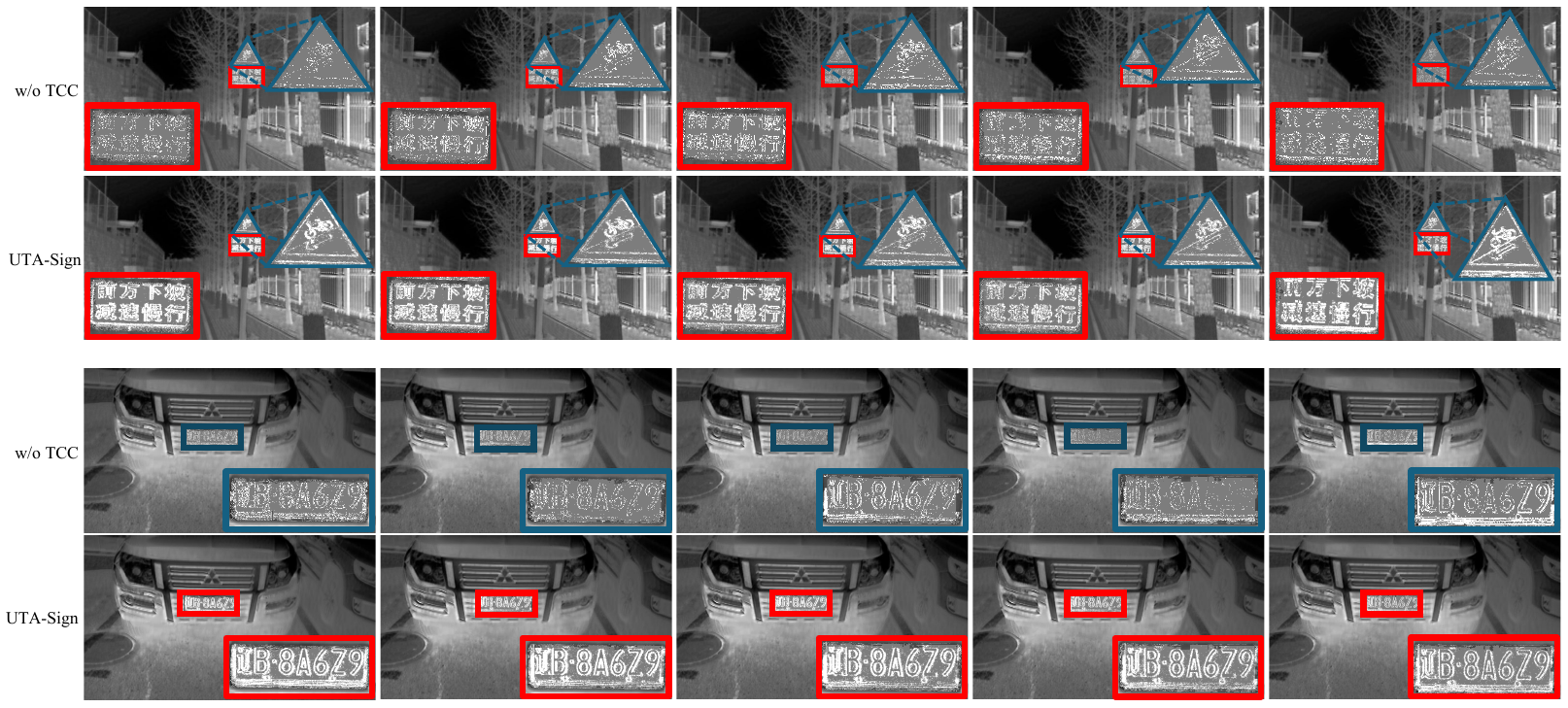}
	\caption{Exemplar results showing the contribution of the temporal consistency constraint module.}
 \vspace{-5mm}
 \label{ablation1_Qualitive}
\end{figure*}

\vspace{2mm}
\noindent \textbf{Contribution from the Temporal Consistency Constraint.\quad}
To evaluate the effectiveness of the TCC module, we conduct ablation experiments comparing the model's performance with and without this module, as shown in Fig.~\ref{ablation1_Qualitive}.

Without the TCC module, the bicycle and arrow symbols in the first scene of Fig. \ref{ablation1_Qualitive} are only faintly recognizable, and the sparse event points do not provide enough information for information perception. In contrast, with TCC module, the multi-frame sketches enhance the reconstruction of the target frame, leading to denser points and clear instruction.

In the second scenario involving license plate numbers, illustrated in Fig.~\ref{ablation1_Qualitive}, the absence of the TCC module results in missing license plate information in the third and fifth reconstructed frames, affecting different regions. However, the TCC module compensates for this missing information by utilizing guidance and constraints from multi-frame data. For example, in the fifth frame, the TCC module draws on adjacent frames to restore the missing details. These results demonstrate that the TCC module effectively maintains temporal visual consistency, which in turn enhances the richness of the semantic information.

\begin{table}[htbp]
  \centering
  \caption{Ablation study of different loss compositions.}
    \begin{tabular}{@{\hskip 4pt} l @{\hskip 6pt} c @{\hskip 6pt} c @{\hskip 6pt} c @{\hskip 6pt} c @{\hskip 6pt} c @{\hskip 4pt}}
    \toprule
    \textbf{Method} & \textbf{NIQE$_\downarrow$} & \textbf{PI$_\downarrow$} & \textbf{EN$_\uparrow$} & \textbf{SD$_\uparrow$} & \textbf{CNNIQA$_\uparrow$} \\
    \midrule
    \textbf{w/o G. Loss} & 5.385  & 4.500  & 7.102  & 45.501  & 0.478  \\
    \textbf{w/o P. Loss} & 5.328  & 4.328  & 7.095  & 44.725  & 0.513  \\
    \textbf{w/o P\&G. Loss} & 5.390  & 4.389  & 7.105  & 45.321  & 0.514  \\
    \textbf{\M } & \textcolor{red}{\textbf{5.229}} & \textcolor{red}{\textbf{4.303}} & \textcolor{red}{\textbf{7.116}} & \textcolor{red}{\textbf{45.590}} & \textcolor{red}{\textbf{0.525}} \\
    \bottomrule
    \end{tabular}%
  \label{tab:ablation}%
\end{table}

\vspace{2mm}
\noindent \textbf{Contributions of Different Loss Function Items.\quad}
We further investigate the contribution of different loss components to the overall performance of the~\M~model. Specifically, we conduct ablation experiments by removing the gradient loss and the perceptual loss, respectively, while keeping other settings unchanged.

As shown in Tab.~\ref{tab:ablation}, the removal of either the gradient loss or the perceptual loss leads to a decline in performance across most metrics. This demonstrates that both losses are essential for enhancing output quality and stability. Interestingly, when both the gradient and perceptual losses are removed simultaneously (referred to as w/o P\&G Loss), certain metrics do not show further deterioration in performance. This can be attributed to the overlapping functions of the two losses, both of which improve edge sharpness and texture details. When only one of the losses is removed, the other can provide some compensation, resulting in a more noticeable drop in performance. In contrast, when both losses are absent, the model training depends entirely on pixel-level loss. This simplifies the objective function and minimizes gradient conflicts that arise from multiple losses, allowing the performance in some metrics to remain close to that observed with the removal of just one loss.

Overall, the full model with all loss components achieves the best quantitative results, confirming the effectiveness of the proposed loss design in improving both low-level image quality and high-level perceptual characteristics.

\section{Conclusion}
This paper presents a novel solution for augmenting thermal video in low-light autonomous driving perception, focusing on traffic signage sketching by utilizing the strengths of event cameras. The proposed method, referred to as \M, constructs a dual-boosting fusion network integrated with an unsupervised training mechanism. This approach targets traffic signage and refines the raw captured data, ensuring robustness and adaptability in various driving scenarios. 
To address the challenges posed by non-uniform and incomplete signals resulting from the asynchronous acquisition of event cameras, this study utilizes contour information derived from thermal imagery to achieve alignment and enhancement of the event signals. Furthermore, the complete event sequence is used to enrich the semantic information of the raw thermal frames. 
Experimental results demonstrate the effectiveness of the proposed method in generating high-quality visual reconstructions and supporting downstream detection tasks. This fusion framework offers a promising solution to the longstanding challenge of reliable signage detection and analysis in extremely low-light driving environments, contributing to advancements in autonomous driving technologies.

\printcredits

\bibliographystyle{elsarticle-num}

\bibliography{cas-refs}

\begin{thebibliography}{10}
\expandafter\ifx\csname url\endcsname\relax
  \def\url#1{\texttt{#1}}\fi
\expandafter\ifx\csname urlprefix\endcsname\relax\def\urlprefix{URL }\fi
\expandafter\ifx\csname href\endcsname\relax
  \def\href#1#2{#2} \def\path#1{#1}\fi

\bibitem{ma2019infrared}
J.~Ma, Y.~Ma, C.~Li, Infrared and visible image fusion methods and applications: A survey, Information Fusion (Elsevier) 45 (2019) 153--178.

\bibitem{DU2025102859}
H.~Du, L.~Ren, Y.~Wang, X.~Cao, C.~Sun, Advancements in perception system with multi-sensor fusion for embodied agents, Information Fusion (Elsevier) 117 (2025) 102859.

\bibitem{GHOSH2025102891}
D.~K. Ghosh, Y.~J. Jung, Depth cue fusion for event-based stereo depth estimation, Information Fusion (Elsevier) 117 (2025) 102891.

\bibitem{bardow2016simultaneous}
P.~Bardow, A.~J. Davison, S.~Leutenegger, Simultaneous optical flow and intensity estimation from an event camera, in: Proceedings of the IEEE Conference on Computer Vision and Pattern Recognition, 2016, pp. 884--892.

\bibitem{scheerlinck2018continuous}
C.~Scheerlinck, N.~Barnes, R.~Mahony, Continuous-time intensity estimation using event cameras, in: Asian Conference on Computer Vision, Springer, 2018, pp. 308--324.

\bibitem{munda2018real}
G.~Munda, C.~Reinbacher, T.~Pock, Real-time intensity-image reconstruction for event cameras using manifold regularisation, International Journal of Computer Vision 126~(12) (2018) 1381--1393.

\bibitem{xu2020fusiondn}
H.~Xu, J.~Ma, Z.~Le, J.~Jiang, X.~Guo, {FusionDN}: A unified densely connected network for image fusion, in: Proceedings of the AAAI Conference on Artificial Intelligence, Vol.~34, 2020, pp. 12484--12491.

\bibitem{ma2022swinfusion}
J.~Ma, L.~Tang, F.~Fan, J.~Huang, X.~Mei, Y.~Ma, {SwinFusion}: Cross-domain long-range learning for general image fusion via swin transformer, IEEE/CAA Journal of Automatica Sinica 9~(7) (2022) 1200--1217.

\bibitem{zhang2021sdnet}
H.~Zhang, J.~Ma, {SDNet}: A versatile squeeze-and-decomposition network for real-time image fusion, International Journal of Computer Vision (2021) 1--25.

\bibitem{LI2024102147}
H.~Li, X.-J. Wu, {CrossFuse}: A novel cross attention mechanism based infrared and visible image fusion approach, Information Fusion (Elsevier) 103 (2024) 102147.

\bibitem{Liang2022ECCV}
P.~Liang, J.~Jiang, X.~Liu, J.~Ma, Fusion from decomposition: A self-supervised decomposition approach for image fusion, in: Proceedings of European Conference on Computer Vision, 2022, pp. 719--735.

\bibitem{wu2022difnet}
M.~Wu, X.~Zhang, X.~Sun, Y.~Zhou, C.~Chen, J.~Gu, X.~Sun, R.~Ji, {DIFNet}: Boosting visual information flow for image captioning, in: Proceedings of the IEEE/CVF Conference on Computer Vision and Pattern Recognition, 2022, pp. 18020--18029.

\bibitem{xu2020u2fusion}
H.~Xu, J.~Ma, J.~Jiang, X.~Guo, H.~Ling, {U2Fusion}: A unified unsupervised image fusion network, IEEE Transactions on Pattern Analysis and Machine Intelligence 44~(1) (2020) 502--518.

\bibitem{9834137}
W.~Tang, F.~He, Y.~Liu, {YDTR}: Infrared and visible image fusion via y-shape dynamic transformer, IEEE Transactions on Multimedia 25 (2023) 5413--5428.

\bibitem{zhang2025daaf}
T.~Zhang, J.~Zhao, Y.~Zhu, G.~Cui, Y.~Jing, Y.~Lyu, {DAAF}: Degradation-aware adaptive fusion framework for robust infrared and visible images fusion, arXiv preprint arXiv:2504.10871 (2025).

\bibitem{yi2024text}
X.~Yi, H.~Xu, H.~Zhang, L.~Tang, J.~Ma, {Text-IF}: Leveraging semantic text guidance for degradation-aware and interactive image fusion, in: Proceedings of the IEEE/CVF Conference on Computer Vision and Pattern Recognition, 2024, pp. 27026--27035.

\bibitem{zhang2025omnifuse}
H.~Zhang, L.~Cao, X.~Zuo, Z.~Shao, J.~Ma, {OmniFuse}: Composite degradation-robust image fusion with language-driven semantics, IEEE Transactions on Pattern Analysis and Machine Intelligence 47~(9) (2025) 7577--7595.

\bibitem{soltangholi2023intensity}
A.~R. Soltangholi, A.~Harati, A.~Vahedian, Intensity-image reconstruction using event camera data by changing in {LSTM} update, in: 2023 13th International Conference on Computer and Knowledge Engineering, IEEE, 2023, pp. 156--161.

\bibitem{lei2024many}
T.~Lei, X.~Guo, Y.~Li, How many events are needed for one reconstructed image using an event camera?, The International Archives of the Photogrammetry, Remote Sensing and Spatial Information Sciences 48 (2024) 645--650.

\bibitem{geng2024event}
M.~Geng, L.~Zhu, L.~Wang, W.~Zhang, R.~Xiong, Y.~Tian, Event-based visible and infrared fusion via multi-task collaboration, in: Proceedings of the IEEE/CVF Conference on Computer Vision and Pattern Recognition, 2024, pp. 26929--26939.

\bibitem{zong2023single}
P.~Zong, Q.~Liu, H.~Deng, Y.~Zhuang, Single pixel event tensor: A new representation method of event stream for image reconstruction, IEEE Sensors Journal 23~(17) (2023) 19590--19597.

\bibitem{liang2023event}
Q.~Liang, X.~Zheng, K.~Huang, Y.~Zhang, J.~Chen, Y.~Tian, Event-diffusion: Event-based image reconstruction and restoration with diffusion models, in: Proceedings of the 31st ACM International Conference on Multimedia, 2023, pp. 3837--3846.

\bibitem{li2024image}
P.~Li, Y.~Zhang, Y.~Fang, An image reconstruction method on brightness constancy for event data, in: 2024 IEEE 6th Advanced Information Management, Communicates, Electronic and Automation Control Conference, Vol.~6, IEEE, 2024, pp. 1512--1516.

\bibitem{guo2023event}
G.~Guo, Y.~Feng, H.~Lv, Y.~Zhao, H.~Liu, G.~Bi, Event-guided image super-resolution reconstruction, Sensors 23~(4) (2023) 2155.

\bibitem{mahmood2024enhanced}
Z.~Mahmood, K.~Khan, M.~Shahzad, A.~Fayyaz, U.~Khan, Enhanced detection and recognition system for vehicles and drivers using multi-scale retinex guided filter and machine learning, Multimedia Tools and Applications 83~(6) (2024) 15785--15824.

\bibitem{9750714}
Z.~Chen, L.~Wang, C.~Wang, Y.~Zheng, Fog image enhancement algorithm based on improved retinex algorithm, in: 2022 3rd International Conference on Electronic Communication and Artificial Intelligence, 2022, pp. 196--199.

\bibitem{yao2023traffic}
J.~Yao, B.~Huang, S.~Yang, X.~Xiang, Z.~Lu, Traffic sign detection and recognition under low illumination, Machine Vision and Applications 34~(5) (2023) 75.

\bibitem{wu2023shadow}
S.~Wu, Z.~Liu, H.~Lu, Y.~Huang, Shadow hunter: Low-illumination object-detection algorithm, Applied Sciences 13~(16) (2023) 9261.

\bibitem{ren2023lightweight}
K.~Ren, Q.~Tao, H.~Han, A lightweight object detection network in low-light conditions based on depthwise separable pyramid network and attention mechanism on embedded platforms, Journal of the Franklin Institute 360~(6) (2023) 4427--4455.

\bibitem{hui2024wsa}
Y.~Hui, J.~Wang, B.~Li, {WSA-YOLO}: Weak-supervised and adaptive object detection in the low-light environment for {YOLOV7}, IEEE Transactions on Instrumentation and Measurement 73 (2024) 1--12.

\bibitem{peng2024novel}
D.~Peng, W.~Ding, T.~Zhen, A novel low light object detection method based on the {YOLOv5} fusion feature enhancement, Scientific Reports 14~(1) (2024) 4486.

\bibitem{vinoth2024lightweight}
K.~Vinoth, P.~Sasikumar, Lightweight object detection in low light: Pixel-wise depth refinement and tensorrt optimization, Results in Engineering 23 (2024) 102510.

\bibitem{lee-2023-edgemultiRGB2TIR}
D.-G. Lee, A.~Kim, Edge-guided multi-domain rgb-to-tir image translation for training vision tasks with challenging labels (2023).

\bibitem{gehrig2020video}
D.~Gehrig, M.~Gehrig, J.~Hidalgo-Carri{\'o}, D.~Scaramuzza, Video to events: Recycling video datasets for event cameras, in: Proceedings of the IEEE/CVF Conference on Computer Vision and Pattern Recognition, 2020, pp. 3586--3595.

\bibitem{10655996}
N.~Zheng, M.~Zhou, J.~Huang, J.~Hou, H.~Li, Y.~Xu, F.~Zhao, Probing synergistic high-order interaction in infrared and visible image fusion, in: 2024 IEEE/CVF Conference on Computer Vision and Pattern Recognition, 2024, pp. 26374--26385.

\bibitem{YI2024102450}
X.~Yi, L.~Tang, H.~Zhang, H.~Xu, J.~Ma, Diff-{IF}: Multi-modality image fusion via diffusion model with fusion knowledge prior, Information Fusion (Elsevier) 110 (2024) 102450.

\bibitem{ma2022toward}
L.~Ma, T.~Ma, R.~Liu, X.~Fan, Z.~Luo, Toward fast, flexible, and robust low-light image enhancement, in: Proceedings of the IEEE/CVF Conference on Computer Vision and Pattern Recognition, 2022, pp. 5637--5646.

\bibitem{cai2023retinexformer}
Y.~Cai, H.~Bian, J.~Lin, H.~Wang, R.~Timofte, Y.~Zhang, Retinexformer: One-stage retinex-based transformer for low-light image enhancement, in: Proceedings of the IEEE/CVF international conference on computer vision, 2023, pp. 12504--12513.

\bibitem{liang2024towards}
G.~Liang, K.~Chen, H.~Li, Y.~Lu, L.~Wang, Towards robust event-guided low-light image enhancement: A large-scale real-world event-image dataset and novel approach, in: Proceedings of the IEEE/CVF Conference on Computer Vision and Pattern Recognition, 2024, pp. 23--33.

\end{thebibliography}



\end{document}